\documentclass[conference]{IEEEtran}
\IEEEoverridecommandlockouts
\usepackage{cite}
\usepackage{amsmath,amssymb,amsfonts}
\usepackage{algorithmic}
\usepackage{graphicx}
\usepackage{textcomp}
\usepackage{xcolor}
\usepackage{subfig}
\usepackage{booktabs}
\usepackage{import}
\usepackage{color}
\usepackage{multirow}
\usepackage{array}
\usepackage[font=footnotesize]{caption}
\usepackage[inline]{enumitem}
\usepackage{makecell}
\usepackage{siunitx}
\usepackage{caption}
\usepackage{soul}
\usepackage{adjustbox}
\usepackage{tikz}
\usepackage{float}
\usetikzlibrary{arrows.meta,positioning,fit,calc}
\begin{document}
\bstctlcite{IEEEexample:BSTcontrol}

%
\title{Collaborative Space Object Detection with Multi-Satellite Viewpoints in LEO Constellations}

\author{
\IEEEauthorblockN{Xingyu Qu\textsuperscript{1}, Wenxuan Zhang\textsuperscript{2}, and Peng Hu\textsuperscript{1,*}}
\IEEEauthorblockA{\textsuperscript{1}Department of Electrical and Computer Engineering, University of Manitoba, Winnipeg, Canada}
\IEEEauthorblockA{\textsuperscript{2}Department of Electrical and Computer Engineering, Queen's University, Kingston, Canada}
\IEEEauthorblockA{*Email: peng.hu@umanitoba.ca}

\thanks{We acknowledge the support provided by the Government of Canada and Natural Sciences and Engineering Research Council of Canada (NSERC), [funding reference number RGPIN-2022-03364].}
}

\maketitle

\begin{abstract}
With the growing number of satellites in low Earth orbit (LEO) constellations, the near-Earth space environment has become increasingly congested, making space object detection (SOD) a pressing challenge for space safety and sustainability. To mitigate collision risks and ensure the continuity of space operations, SOD systems must deliver fast and accurate detection under stringent onboard constraints. In this paper, we investigate the potential of multi-viewpoint observation fusion within a deep learning (DL) framework to enhance SOD performance. We design a practical multi-view pipeline and several input representations for feeding multi-view data into YOLO-based detectors. Our experiments show that using multi-view inputs is feasible in most cases and typically produces better results for mAP50 and mAP50-95. For example, in model YOLOv9-m, single-view compared to a three-view fused RGB setting, mAP50 increases from 0.638 to 0.732, while mAP50-95 improves from 0.227 to 0.276. Compared with the single-view setting, the best three-view grayscale configuration improves mAP50 by 36.3\% and mAP50-95 by 46.5\%. These findings establish multi-view fusion as a viable and effective strategy for SOD, with broad implications for space situational awareness in LEO constellation deployments.

\end{abstract}


%
\IEEEpeerreviewmaketitle

\section{Introduction}
%
%
%
%
\IEEEPARstart{W}{ith} the rapid expansion of human activities in outer space, the orbital environment, especially low earth orbit (LEO), has become increasingly overcrowded, and this trend is expected to continue in the coming years.  
As the number of satellites continues to rise, tasks such as conjunction assessment and collision avoidance will become more challenging and demanding \cite{ravi2025comparative}. In a crowded orbital environment, satellites must be able to quickly identify nearby objects to maintain space situational awareness (SSA) and ensure safe operations. This trend highlights the importance of effective space object detection (SOD). However, SOD is challenging because the detection process must meet both high precision and low latency requirements\cite{sensing_vit}. Automated and scalable collision-avoidance methods have attracted increasing attention from researchers. \cite{ravi2025comparative}. Because satellites often have to observe targets from far away and against a dark background, these targets can be quite small and difficult to detect. At the same time, onboard systems must operate under actionable power, hardware, and sensing constraints \cite{ai_driven_collab}.

Under these limitations, relying on a single observation may not fully exploit the information available for strong detection. Multiple different satellites in a cluster observe the same target from different viewpoints, and through information sharing, potentially provide complementary spatial information beyond a single-view \cite{ai_driven_collab}. Meanwhile, deep learning (DL) provides a powerful framework to automatically extract meaningful features from visual inputs. This reduces reliance on manual analysis and improves detection efficiency in increasingly complex orbital environments. 

To address the aforementioned challenges, this paper develops a practical DL-based multi-view pipeline for SOD. The main contributions of this work are summarized as follows:

\begin{itemize}
    \item We propose a lightweight and practical multi-view input pipeline that incorporates multi-view information into a single detector. 
    
    \item We demonstrate that multi-view SOD can be achieved under both RGB and grayscale input settings, indicating that the multi-view strategy is applicable to various visual representations.
    
    \item Our results show that appropriate multi-view inputs improve both detection accuracy and localization quality, as seen in mAP50 and mAP50-95, while keeping inference time and communication costs practical.

\end{itemize}

The remainder of this paper is structured as follows. Section II reviews recent developments in SOD and related studies on multi-view. Section III presents the system model and DL pipeline. Section IV presents the experimental setup and performance evaluation results. Section V concludes the paper and outlines the future work.

\section{Related Work}

In the current development of SOD, one major problem is the lack of high-quality, labeled, and publicly available space-image datasets. To overcome these limitations, researchers constructed simulated space-target datasets designed to capture a range of orbital scenarios, sensing conditions, and target poses. The availability of such resources enabled the application of DL to SOD and shifted research attention toward the design and optimization of detection frameworks~\cite{spark,spacecraft_dataset}. At the same time, SOD systems are deployed in space environments where computational resources, and size, weight, and power (SWaP) footprint are limited, making it necessary to seek detection methods that maintain reliable performance within efficiency budgets~\cite{toward_onboard_ai}. 

In particular, convolutional neural network (CNN)-based and You Only Look Once (YOLO)-style models have gained attention because they provide a positive balance between detection accuracy and efficiency~\cite{toward_onboard_ai}. The core advantage of the YOLO series is using a unified and efficient one-stage framework to achieve real-time object detection\cite{yolo_original}. As development has progressed, YOLO models have continued to improve in both detection precision and computational efficiency. For instance, YOLOv2 introduced anchor-based prediction and improved training strategies, resulting in significantly higher accuracy for real-time detection tasks~\cite{redmon2017yolo9000}. YOLOv3 further advanced detection performance through a more powerful backbone and multi-scale prediction~\cite{redmon2018yolov3}. With YOLOv7, the architectural design and training methodologies were further redesigned, leading to a significant improvement in the trade-off between detection speed and accuracy~\cite{wang2023yolov7}.

Building on these developments, YOLOv9 provides a strong baseline for SOD. It integrates two important components, named Programmable Gradient Information (PGI) and the Generalized Efficient Layer Aggregation Network (GELAN)\cite{yolov9}. These designs are especially relevant to SOD, where targets often occupy only a very small number of pixels and images are marked by dark backgrounds and low resolution. Under such conditions, useful information can be gradually compressed or lost during convolutional processing.  PGI helps address this problem by reducing the loss of deep-level information, enabling the model to learn more targets.  In addition, GELAN improves parameter utilization and feature aggregation while maintaining a lightweight architecture, reducing input redundancy introduced by the multi-view method\cite{yolov9}. 

However, despite these advantages, the YOLO series is designed by default to accept single images with three channels as input, meeting the requirements for single-view scenarios and providing a baseline. This also indicates that YOLOv9 is originally intended for single-input settings and cannot directly learn complementary information from multi-view inputs. This limitation is important in SOD, where improvements to the detector may not be the most efficient solution. Existing research has focused on architectural modifications such as attention mechanisms and transformer-based modeling to improve target detection \cite{small_object_survey}. Although these methods can improve performance, they often require more complex detector designs and higher computational costs, which may limit their applicability in space environments.

Meanwhile, multi-view research has demonstrated that different perspectives of the same object can be used as joint inputs, and better representations can be learned by modeling the relationships between these perspectives\cite{li2018survey}. By combining these perspectives, it is possible to construct more comprehensive data representations than single-view approaches \cite{li2018survey}. In space target observation scenarios, spacecraft formation and dual-point-of-view (DPOV) schemes have further demonstrated the practical feasibility of acquiring multiple observations of the same target \cite{space_dpov}. These studies suggest that multi-view information can provide complementary cues and that multi-view observation is available in space applications. However, existing multi-view research does not specifically address downstream SOD detection tasks. Studies on multi-view observation in space primarily focus on observation geometry and orbit-related analysis, rather than on end-to-end visual detection. As a result, there is still limited research on how to incorporate complementary multi-view information into the training of visual detection models for SOD.

Motivated by this gap, our work explores a multi-view input design for SOD based on early fusion. To enable YOLOv9 to process multi-view observations, information from different views is integrated into a unified input representation before being fed into the detector. In this way, the same YOLOv9 architecture can be used for both single-view and multi-view scenarios. Compared with introducing additional multi-branch structures or more complex fusion modules, this design helps control model complexity and supports fair experimental comparisons. Therefore, the study can focus directly on how multi-view input representations affect SOD performance.

\section{System model}

\begin{figure*}[t]
  \centering
  \includegraphics[width=0.80\textwidth,trim=0 45 0 0,clip]{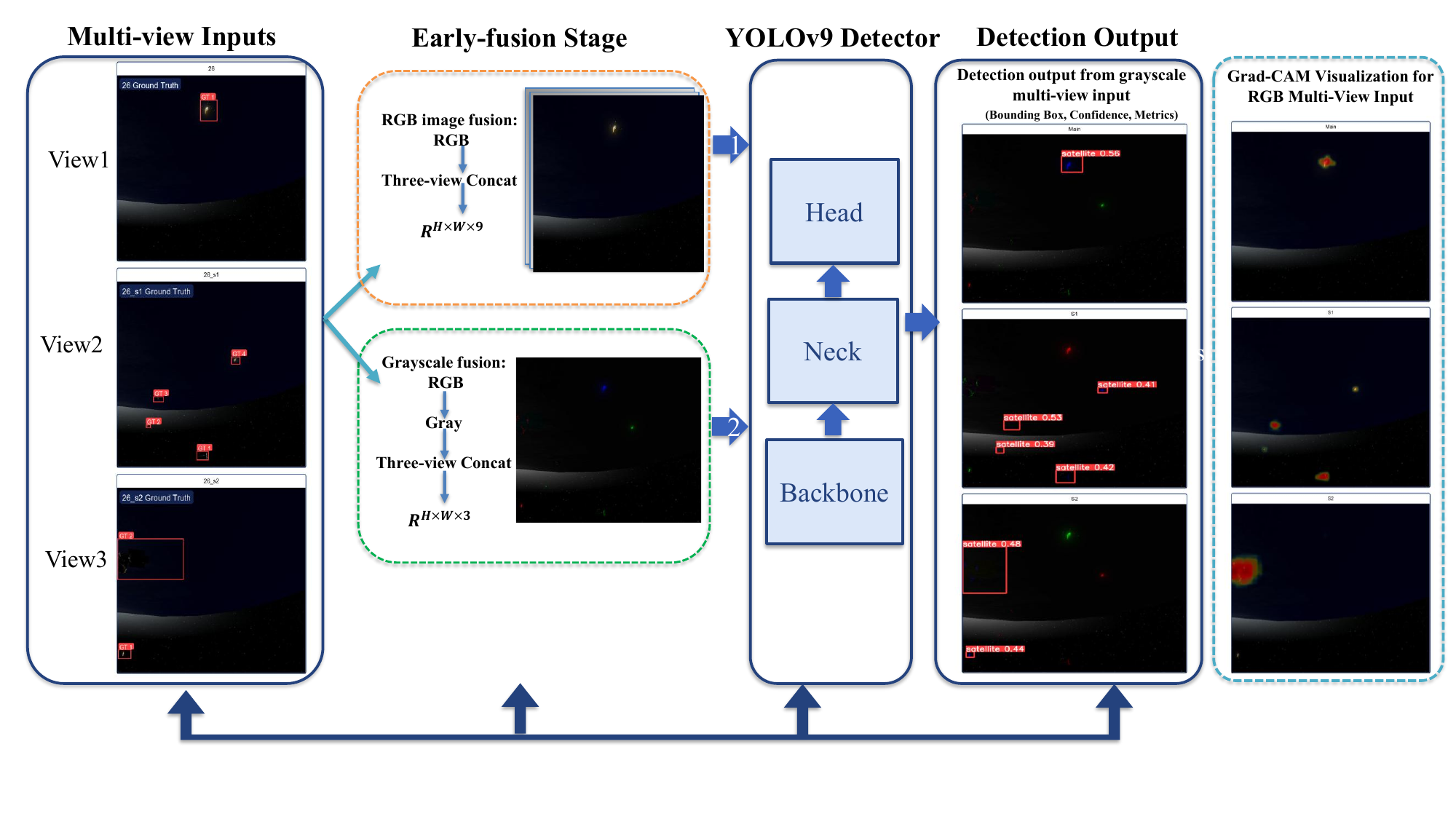}
  \caption{An illustration of the proposed multi-view early-fusion pipeline. The detection output column shows the raw detection results from the fused grayscale multi-view input. On the right, the Grad-CAM visualizations show model attention, based on detector outputs generated from the fused RGB multi-view input.}
  \label{fig:mv_pipeline}
\end{figure*}


\subsection{Multi-View SOD}
In this study, SOD is defined under both single-view and multi-view settings. For the single-view baseline, each image is treated as an independent sample. The single-view input is expressed as $X_{\text{single-view}} = \{X_n\}_{n=1}^{N}$. Where $X_n$ denotes the $n$-th image sample, and $N$ is the total number of samples. In this setting, the detector could only learn features from a single observation. To provide comparison with the baseline, we introduced a multi-view input: $X_{\text{multi-view}} = \{(X_{j,1}, X_{j,2}, \ldots, X_{j,k})\}_{j=1}^{N}$. Here, $X_{j,1}$, $X_{j,2}$, $\ldots$, $X_{j,k}$ represent the $N$ clusters containing the number of $k$ viewpoint. In this study, the primary view represents the view in a fused multi-view sample whose annotations are used as the supervision target during training and evaluation. The support views represent the remaining views in the fused sample that provide complementary visual information, but whose annotations are not used as supervision targets for that sample.

\subsection{Satellite Formation in Multi-View SOD }
To support this multi-view exploration, we consider a satellite formation in which one satellite view $X_{j,1}$ is used as the central observation while the other satellite views act as auxiliary observations. The SOD Clustering dataset used in this work organizes each sample as a satellite cluster with multiple viewpoints. 
The $i$-th satellite cluster is denoted as $S_i = \{s_{i,1}, s_{i,2}, \ldots, s_{i,k}\}$,
where $k$ is the number of satellites in the $i$-th cluster\cite{ai_driven_collab}.
Here, $s_{i,1}$ denotes the center satellite, while $s_{i,2}, \ldots, s_{i,k}$ denote the support satellites. The support satellites are positioned within a predefined radius $r_i$ of the center satellite $s_{i,1}$. In this way, the formation consists of one center satellite and multiple nearby support satellites distributed around it within a controlled spatial range\cite{ai_driven_collab}.

This design of satellite formation ensures that the same target scene is observed from multiple satellites. The corresponding views of the $i$-th cluster are represented as $V_i = \{V_{i,1}, V_{i,2}, \ldots, V_{i,k}\}$, where $V_{i,1}$ is captured by the center satellite $s_{i,1}$, and $V_{i,2}, \ldots, V_{i,k}$ are captured by the support satellites $s_{i,2}, \ldots, s_{i,k}$, respectively \cite{ai_driven_collab}. Accordingly, the view $V_{i,1}$ is defined as the center view, and $V_{i,2}, \ldots, V_{i,k}$ are denoted as the support view. Therefore, this design enables satellites to observe the same scene from different viewing angles and at different distances.


\subsection{Proposed DL Pipeline}

This subsection describes how the original images are organized into suitable training samples for the detector. In this study, early fusion means combining multiple views at the input stage by concatenating them into a single fused sample before feature extraction and detection. We use this strategy mainly because it is easy to implement and does not require major changes to the original detector architecture. It also helps keep the model relatively lightweight, with limited extra computational cost and inference overhead. Another practical advantage is that the fused inputs can be built directly from the existing multi-view dataset, without the need for additional manual annotation.

As described earlier, the multi-view input within each cluster is denoted as $X_j^{\text{multi-view}} = (X_{j,1}, X_{j,2}, \ldots, X_{j,k})$, where $X_{j,i}$ represents the input image associated with view $V_{j,i}$. To construct a multi-view input from $k$ RGB images, the $k$ views are concatenated along the channel dimension to form a $3k$-channel feature map. If each view is an RGB image of size $H \times W \times 3$, the fused input is represented as $X_j^{\text{fusion}} = \mathrm{Concat}(X_{j,1}, X_{j,2}, \ldots, X_{j,k}) \in \mathbb{R}^{H \times W \times 3k}$. Using this strategy, each fused input is supervised using only the annotations of its primary view during training. In this study, the primary view is the first view in the current arrangement, and its relevant label is used for supervision during training.


We used a repeated stack strategy, where each view in the $j$-th cluster is treated as the primary view in turn. For the $m$-th arrangement, the input captured by view $V_{j,m}$ is placed first and treated as the primary view input. The remaining inputs are arranged sequentially after the primary view. In this arrangement, $X_{j,m}$ is denoted as the primary view input, and the label associated with $X_{j,m}$ is used as the supervision information, while all remaining inputs are treated as support view inputs. The fused input from this arrangement is defined as $X_j^{(m)} = \mathrm{Concat}(X_{j,m}, X_{j,m+1}, \ldots, X_{j,k}, X_{j,1}, \ldots, X_{j,m-1})$, where $m \in \{1,2,\ldots,k\}$. By repeating this process for all $m \in \{1,2,\ldots,k\}$, every view in the cluster is treated once as the primary view, and the labels of each view are included in training.


Grayscale images can further evaluate the effectiveness of multi-view inputs without increasing the complexity of the model. A grayscale image is obtained from the original RGB image using the standard luminance transformation: $I_{gray} = 0.299R + 0.587G + 0.114B$.
Accordingly, the grayscale version of the $i$-th view in the $j$-th cluster is defined as $G_{j,i} = I_{\mathrm{gray}}(X_{j,i})$, where $G_{j,i}$ denotes the grayscale conversion result of $X_{j,i}$.


For the multi-view grayscale setting, the $k$ grayscale views are concatenated following the same strategy as in the RGB-based multi-view setting. Let the grayscale views of the $j$-th cluster be denoted as $(G_{j,1}, G_{j,2}, \ldots, G_{j,k})$. These grayscale images are also concatenated along the channel dimension: $X_j^{\mathrm{gray}} = \mathrm{Concat}(G_{j,1}, G_{j,2}, \ldots, G_{j,k})$, where $X_j^{\mathrm{gray}} \in \mathbb{R}^{H \times W \times k}$. In addition, the same repeated stacking strategy is applied in the grayscale setting. Each view is first converted to grayscale, and then each grayscale view is used once as the primary view while the remaining grayscale views perform as support views. For the $m$-th arrangement, the fused grayscale input is defined as $X_{j,\text{gray}}^{(m)} = \mathrm{Concat}(G_{j,m}, G_{j,m+1}, \ldots, G_{j,k}, G_{j,1}, \ldots, G_{j,m-1})$, where $m \in \{1,2,\ldots,k\}$.


Figure~\ref{fig:mv_pipeline} shows the end-to-end multi-view early-fusion pipeline used for the SOD\_clustering dataset\cite{ai_driven_collab}. In this dataset, each satellite cluster consists of a center satellite and two support satellites. The $i$-th cluster can be written as $S_i = \{s_{i,1}, s_{i,2}, s_{i,3}\}$, with its corresponding views defined as $V_i = \{V_{i,1}, V_{i,2}, V_{i,3}\}$. Using this three-view setup, we prepare the inputs in two different ways. In the RGB version, the three RGB views are concatenated along the channel dimension, creating a fused tensor in $\mathbb{R}^{H \times W \times 9}$. In the grayscale version, each view is first converted to grayscale, and the three grayscale views are then concatenated to form a fused tensor in $\mathbb{R}^{H \times W \times 3}$. The output after RGB fusion is not a standard image. Instead, it is better understood as a 9-channel feature map formed by combining the three RGB views. Also, the fused grayscale result is a 3-channel tensor, which can be stored in PNG format to avoid compression loss. However, this does not mean that the resulting file is a natural RGB image. Each channel represents a grayscale observation from a different view.

For training, we also apply a repeated stacking strategy. This means that each view in a cluster takes a turn serving as the primary view, while the other two act as support views. The fused inputs produced by these two processing methods are then fed separately into the detector for training and evaluation. As mentioned earlier, the fused RGB input is not a standard image but a multi-channel representation of multiple views. The detector doesn't produce a visual output from the fused input itself. Because of this, the detector performance of fused RGB input is evaluated mainly through metrics such as mAP50 and mAP50-95. To improve interpretability, we further apply Grad-CAM++ \cite{chattopadhay2018grad}. This method highlights the regions with the greatest influence on the model's predictions and produces a heatmap. When the heatmap is overlaid on the original image, it provides a clearer visual explanation of what the model is focusing on and helps us assess the detection results more effectively.




\subsection{Proposed Model Architecture}

The model architecture in this experiment is mainly built on the YOLO framework. The key difference is that we implement a multi-view input representation that can be embedded into YOLO, allowing multi-view information to be learned using only a single detector. For RGB images, the three views are concatenated into a 9-channel tensor. To handle this setting, the first convolutional layer is modified from 3 input channels to 9 input channels. In contrast, after concatenation, the multi-view grayscale images still satisfy the default 3-channel input format of the YOLO framework, and can be directly fed for model training. As a result, the modification is limited to the input preprocessing stage, while the downstream YOLO backbone, neck, and detection head remain unchanged. This design enables a fair comparison between single-view and multi-view settings while minimizing changes to the original detector architecture.

\section{Experimental Results}

\subsection{Experiment Setup}
All experiments in this study are operated on the Nibi HPC cluster. The model implementation is based on the YOLOv9 framework, using Python 3.11.14, PyTorch 2.2.2, and CUDA 12.1. The training process is performed on an NVIDIA H100 80GB HBM3 GPU with eight worker threads for data loading.

The SOD\_Clustering dataset is split into 80\% training data and 20\% validation data. This split is operated before the early-fusion preprocessing. The dataset contains 180 images in total. All images were resized to $640 \times 640$. The batch size is fixed at 16 throughout all experiments.

Although the proposed framework supports a general multi-view setting with an arbitrary number of views, the SOD\_Clustering dataset\cite{ai_driven_collab} used in this study provides three-view samples, and therefore all experiments are conducted under a three-view configuration. In the single-view input setting, the images from the dataset are treated as independent samples and fed into the YOlOv9 detector. To evaluate the performance of the multi-view approach in SOD, a single-view RGB input and a single-view grayscale input are used as baselines. For comparison with the baseline, we consider two proposed multi-view input types: RGB fusion and grayscale fusion. In addition, four detector settings are evaluated: YOLOv9-t with COCO pretrained weights, YOLOv9-t without COCO pretrained weights, YOLOv9-m without pretrained, and GELAN-t without pretrained. All models are trained for up to 1000 epochs using early stopping with a patience of 300 epochs.

\begin{table*}[t]
\vspace{6pt}
\centering
\caption{Comparison of single-view (SV) and three-view (3V) original-image inputs under the same model.}
\label{tab:original_sv_vs_3v}
\footnotesize
\setlength{\tabcolsep}{3pt}
\renewcommand{\arraystretch}{0.95}
\begin{tabular}{llccccccccc}
\toprule
& & \multicolumn{3}{c}{mAP50} & \multicolumn{3}{c}{mAP50-95} & \multicolumn{3}{c}{Inf. Time} \\
\cline{3-11}
Model & Setting & SV & 3V & $\Delta$ & SV & 3V & $\Delta$ & SV & 3V & $\Delta$ \\
\hline
YOLOv9-t & w/o pretrain & 0.686 & 0.673 & -0.013 (-1.9\%) & 0.228 & 0.263 & +0.035 (+15.4\%) & 10.3 & 15.1 & +4.8 (+46.6\%) \\
YOLOv9-t & w/ COCO pretrain & 0.668 & 0.758 & +0.090 (+13.5\%) & 0.276 & 0.321 & +0.045 (+16.3\%) & 10.9 & 14.0 & +3.1 (+28.4\%) \\
YOLOv9-m & w/o pretrain & 0.638 & 0.732 & +0.094 (+14.7\%) & 0.227 & 0.276 & +0.049 (+21.6\%) & 18.8 & 18.5 & -0.3 (-1.6\%) \\
GELAN-t & w/o pretrain & 0.709 & 0.755 & +0.046 (+6.5\%) & 0.223 & 0.266 & +0.043 (+19.3\%) & 9.0 & 14.1 & +5.1 (+56.7\%) \\
\bottomrule
\end{tabular}
\vspace{4pt}
\parbox{0.82\textwidth}{ $\Delta$ denotes the relative percentage change with respect to the single-view result.}
\end{table*}



Two main evaluation metrics, mAP50 and mAP50-95, are used to determine the performance of multi-view input. We evaluate the practical feasibility of the proposed multi-view setting by inference time and communication cost. In this work, inference time is defined as the average processing time during the validation stage per image. The communication cost is measured in terms of latency. For the \(j\)-th multi-view sample, the total communication latency is defined as the sum of the propagation delay and the transmission delay, given by \(L_j = \frac{d_{m,s_1}+d_{m,s_2}}{c}+\frac{S_{s_1}+S_{s_2}}{b}\). Where \(d_{m,s_1}\) and \(d_{m,s_2}\) denote the distances from the \textit{main} satellite to the two \textit{support} satellites, \(c\) is the signal propagation speed, \(S_{s_1}\) and \(S_{s_2}\) are the image sizes from the two support satellites, and \(b\) denotes the inter-satellite link capacity. 

\subsection{Comparison of Single-View and Three-View RGB Inputs}

As shown in Table~\ref{tab:original_sv_vs_3v}, all input representations are based on the original images without grayscale conversion. Overall, the three-view inputs outperform the single-view input in most settings for both mAP50 and mAP50-95. The main exception is YOLOv9-t without COCO pretrained weights, where the single-view and three-view settings achieved very similar mAP50 scores. However, the three-view input still gained a notable improvement of approximately \(15.4\%\) in the stricter mAP50-95 metric. This result suggests that the three-view strategy helped the model to learn more accurate spatial representation, resulting in more precise bounding boxes.

When COCO pretrained weights are used, the three-view input improved mAP50 by \(13.5\%\) and mAP50-95 by \(16.3\%\) for YOLOv9-t. This indicates that pretraining provided more general visual features, such as edges, textures, and object shapes, which help the model extract and integrate the additional multi-view information more effectively and reliably.

The largest improvement comes from using YOLOv9-m without pre-trained weights. With this configuration, the three-view learning model’s mAP50 increased from 0.638 to 0.732, while mAP50-95 rise from 0.227 to 0.276. This implies that larger model capacity is better suited for more complex multi-view learning and lead to more marked improvements.

In terms of inference time, single-view input is faster than three-view input for YOLOv9-t (with and without COCO pretrained weights) and GELAN-t, with a typical gap of \(3\text{--}5\) ms. In contrast, for YOLOv9-m, the detection speed remains nearly unchanged between the single-view and three-view settings. Three-view inputs require more computational resources, which is expected because they contain more information than single-view inputs. For lightweight models such as YOLOv9-t and GELAN-t, where the baseline inference time is relatively short, this added cost becomes more noticeable. In contrast, YOLOv9-m already involves more parameters and convolutional layers, so the extra time required to process the three-view input is relatively short and therefore appears nearly unchanged.

\begin{table*}[t]
\vspace{6pt}
\centering
\caption{Comparison of single-view (SV) and three-view (3V) grayscale inputs under the same model.}
\label{tab:gray_sv_vs_3v}
\footnotesize
\setlength{\tabcolsep}{3pt}
\renewcommand{\arraystretch}{0.95}
\begin{tabular}{llccccccccc}
\toprule
& & \multicolumn{3}{c}{mAP50} & \multicolumn{3}{c}{mAP50-95} & \multicolumn{3}{c}{Inf. Time} \\
\cline{3-11}
Model & Setting & SV & 3V & $\Delta$ & SV & 3V & $\Delta$ & SV & 3V & $\Delta$ \\
\hline
YOLOv9-t & w/o pretrain & 0.590 & 0.804 & +0.214 (+36.3\%) & 0.200 & 0.293 & +0.093 (+46.5\%) & 9.9 & 9.8 & -0.1 (-1.0\%) \\
YOLOv9-t & w/ COCO pretrain & 0.701 & 0.731 & +0.030 (+4.3\%) & 0.236 & 0.286 & +0.050 (+21.2\%) & 10.9 & 10.3 & -0.6 (-5.5\%) \\
YOLOv9-m & w/o pretrain & 0.589 & 0.762 & +0.173 (+29.4\%) & 0.191 & 0.288 & +0.097 (+50.8\%) & 17.5 & 17.0 & -0.5 (-2.9\%) \\
GELAN-t & w/o pretrain & 0.654 & 0.708 & +0.054 (+8.3\%) & 0.214 & 0.254 & +0.040 (+18.7\%) & 10.1 & 9.2 & -0.9 (-8.9\%) \\
\bottomrule
\end{tabular}
\vspace{4pt}
\parbox{0.82\textwidth}{$\Delta$ denotes the relative percentage change with respect to the single-view result.}
\end{table*}

\subsection{Comparison of Single-View and Three-View Grayscale Inputs}

Table~\ref{tab:gray_sv_vs_3v} mainly compares single-view and three-view grayscale inputs with the same parameter settings. In summary, three-view grayscale input brings a huge increase relative to single-view grayscale inputs. The best-performing model is YOLOv9t without pretrained weights, showing a 36.3\% improvement in mAP50 and a larger 46.5\% improvement in mAP50-95. This result matches our observations using fused RGB inputs. When the inputs are converted to grayscale, the complementary information provided by multiple perspectives becomes more pronounced. Using this support information, the model learns more detailed satellite shapes and target boundaries, which enhances detection performance and generates more accurate bounding boxes.

For the larger YOLOv9-m model, the relative improvement in mAP50-95 (+51.6\%) is the largest among all settings, suggesting that higher-capacity models better capture multi-view information under stricter localization metrics. In contrast, the lightweight GELAN-t model still benefits from a three-view input, with improvements of +8.3\% in mAP50 and +18.7\% in mAP50-95, but its final mAP50-95 remains lower at 0.254. This suggests that lightweight models cannot capture as many features from grayscale multi-view inputs, and their best possible performance is ultimately restricted compared to higher-capacity models.

Comparing the $\Delta$ Relative Percentage Change values between the grayscale and RGB settings in Tables~\ref{tab:original_sv_vs_3v} and ~\ref{tab:gray_sv_vs_3v}, we observe that the grayscale setting generally shows a better performance gain under the multi-view design. For YOLOv9-m, the relative improvement brought by multi-view input increases from 14.7\% in the RGB setting to 29.4\% in the grayscale setting for mAP50, and from 21.6\% to 50.8\% for mAP50-95. A similar result is observed for YOLOv9-t without pretrained weights, where mAP50 improves from -1.9\% to +36.3\%, and mAP50-95 improves from +15.4\% to +46.5\%. This indicates that the benefit of multi-view input becomes more pronounced when color and appearance information is reduced. In these cases, the detector has to rely more on complementary shape and spatial features across different viewpoints, leading to a larger performance gain than in the RGB setting.


Meanwhile, the inference time remained very similar between the two settings, although the three-view grayscale input uses 3 channels and the single-view grayscale input has only 1 channel. A possible explanation is that this difference only affects the first convolutional layer, while most of the computation is still concentrated in the backbone and detection head. As a result, the overall inference time difference between the two input settings remains small.

These results help clarify why multi-view input is effective for SOD. Much of the improvement appears in the stricter mAP50-95 metric, showing that multi-view input primarily helps the model localize objects more precisely instead of just focusing on whether the model can find the object. The advantage of multi-view input becomes more obvious when color information is limited. As seen in the comparison between RGB and grayscale images, the model relies more on structural and spatial details across different views. Multi-view observations also provide the model with extra context, helping reduce uncertainty and missed detections compared to relying on a single observation. It is also worth noting that models with more capacity tend to benefit more from multi-view input. These models can make better use of the extra information available from different viewpoints, leading to larger improvements in detection performance.

\subsection{Communication Cost of Multi-View Solution}
\begin{figure}[t]
    \centering
    \includegraphics[width=0.98\linewidth]{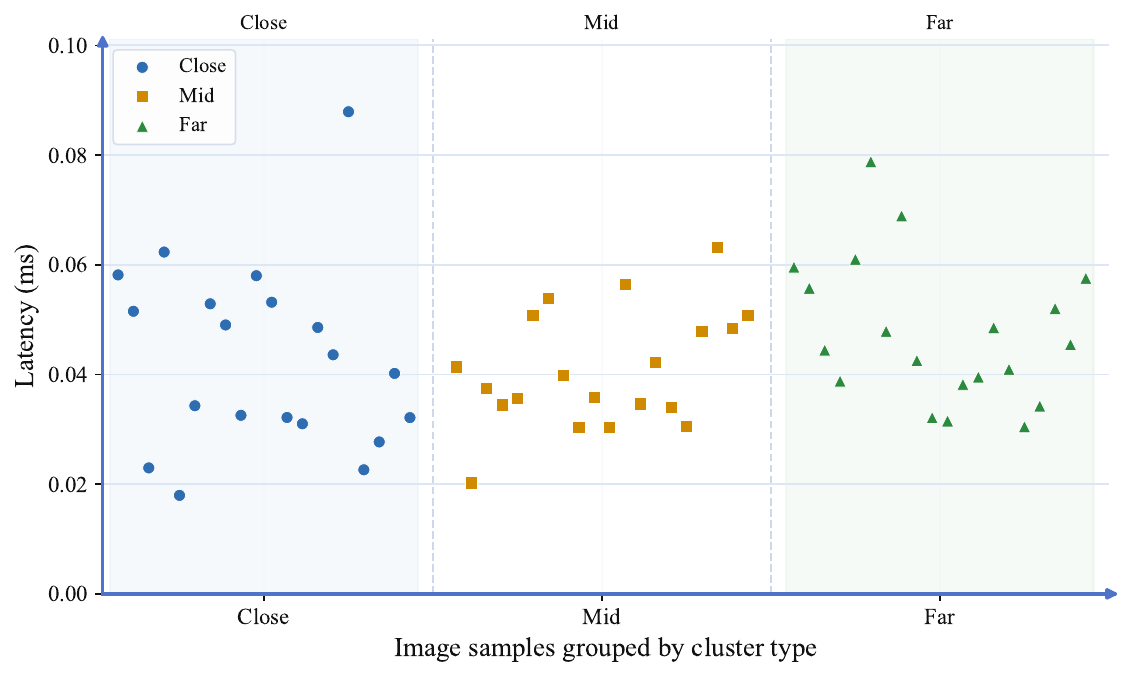}
    \caption{Communication cost (including the components in propagation and transmission delay considering link capacity $b$=10 Gbps) between main and support satellites for viewpoint image transmission versus image indices}
    \label{fig:communication_cost}\vspace{-8pt}
\end{figure}

The proposed multi-view solution implies transferring viewpoint images from support satellites to the main satellite. Here we evaluate the cost of such a transmission process required. As shown in Fig. \ref{fig:communication_cost}, the x-axis represents the grouped image indices under the close, mid, and far distance settings, while the y-axis represents the corresponding communication latency. It is important to note that the close, mid, and far labels refer to the distance between the satellite and the observed target, rather than the inter-satellite distance. The latency values range approximately from 0.02 ms to 0.08 ms, indicating the communication cost of multi-view image transmission is very small. This is mainly due to the reasonable image resolution we used and the high capacity of free-space optical crosslinks on modern satellites.

\section{Conclusion}
This paper addresses the fundamental question of how to efficiently use multi-view input presentations for SOD. We propose a simple and practical pipeline that allows the detector to use multi-view information. The experimental results confirm that multi-view detection effectively improves detection performance in terms of mAP50 and mAP50-95. We also analyze inference time and communication cost to provide a more comprehensive evaluation of the practicality of the proposed method. Overall, our study shows the proposed multi-view approach is a feasible and effective method for enhancing SOD under controlled settings, while laying groundwork for employing satellite constellations in SOD and SSA. Our future work will focus on expanding datasets to incorporate additional spacecraft and exploring alternative multi-view fusion strategies. 


%





\ifCLASSOPTIONcaptionsoff
  \newpage
\fi



\bibliographystyle{IEEEtran}
\bibliography{references}
%



%





\end{document}